\documentclass[lettersize,journal]{IEEEtran}
\usepackage{amsmath,amsfonts}
\usepackage{algorithmic}
\usepackage{algorithm}
\usepackage{array}
\usepackage[caption=false,font=footnotesize,labelfont=rm,textfont=rm]{subfig}
\usepackage{textcomp}
\usepackage{stfloats}
\usepackage{url}
\usepackage{verbatim}
\usepackage{graphicx}
\usepackage{cite}
\usepackage{graphicx}
\usepackage{xcolor}
\ifodd 0
\newcommand{\zyxrev}[1]{{\color{purple}#1}}
\else
\newcommand{\zyxrev}[1]{#1} 
\fi

\begin{document}

\title{SL-ACC: A Communication-Efficient Split Learning Framework with Adaptive Channel-wise Compression}

\author{Zehang Lin, Zheng Lin, Miao Yang, Jianhao Huang, Yuxin Zhang, Zihan Fang, Xia Du, Zhe Chen, Shunzhi Zhu, and Wei Ni,~\IEEEmembership{Fellow,~IEEE}
\thanks{Z. Lin, M. Yang, X. Du, and S. Zhu are with the School of Computer and Information
 Engineering, Xiamen University of Technology, Xiamen, 361000, China (email: 2322071023@stu.xmut.edu.cn; yangmiao@xmut.edu.cn; duxia@xmut.edu.cn; szzhu@xmut.edu.cn). }
 \thanks{Z. Lin, J. Huang are with the Department of Electrical and Electronic Engineering, The University of Hong Kong, Hong Kong. (e-mail: linzheng@eee.hku.hk; jianhaoh@hku.hk)}
\thanks{Y. Zhang, Z. Fang, and Z. Chen are with the Institute of Space Internet, Fudan University, Shanghai 200438, China (e-mail: yuxinzhang22@m.fudan.edu.cn; fang19@fudan.edu.cn; zhechen@fudan.edu.cn).}
\thanks{W. Ni is with Data61, CSIRO, Marsfield, NSW 2122, Australia, and the School of Computing Science and Engineering, the University of New South Wales, Kensington, NSW 2052, Australia (e-mail:
wei.ni@ieee.org).}
}

\markboth{}%
{Shell \MakeLowercase{\textit{et al.}}: A Sample Article Using IEEEtran.cls for IEEE Journals}

\maketitle

\begin{abstract}
The increasing complexity of neural networks poses a significant barrier to the deployment of distributed machine learning (ML) on resource-constrained devices, such as federated learning (FL). Split learning (SL) offers a promising solution by offloading the primary computing load from edge devices to a server via model partitioning. However, as the number of participating devices increases, the transmission of excessive smashed data (i.e., activations and gradients) becomes a major bottleneck for SL, slowing down the model training. To tackle this challenge, we propose a communication-efficient SL framework, named SL-ACC, which comprises two key components: adaptive channel importance identification (ACII) and channel grouping compression (CGC). ACII first identifies the contribution of each channel in the smashed data to model training using Shannon entropy. Following this, CGC groups the channels based on their entropy and performs group-wise adaptive compression to shrink the transmission volume without compromising training accuracy. Extensive experiments across various datasets validate that our proposed SL-ACC framework takes considerably less time to achieve a target accuracy than state-of-the-art benchmarks.
\end{abstract}


\section{Introduction}\label{sec_intro}
\vspace{-0.2em}
\IEEEPARstart{W}{ith} the proliferation of Internet of Things (IoT) devices, massive volumes of data are being generated at the network edge. According to the International Data Corporation (IDC), global data generation is expected to reach approximately 221 zettabytes by 2026~\cite{rydning2023worldwide}. This unprecedented amount of data significantly advances the maturity of machine learning (ML), leading to breakthrough successes across various domains such as autonomous driving~\cite{lin2022channel,wang2018networking,lin2022tracking}, smart healthcare~\cite{fang2025dynamic,tang2024merit,fang2024ic3m}, and natural language processing~\cite{qiu2024ifvit,lin2025hsplitlora,sai2024dmdat}.

The status quo of ML frameworks predominantly relies on a centralized learning (CL) paradigm, where raw data is collected and processed at a central server. However, CL necessitates the transmission of raw data from all participating devices to a server, raising serious privacy risks. Moreover, transmitting large volumes of raw data incurs substantial communication overhead, rendering the deployment of CL infeasible. Federated learning (FL)~\cite{chen2025mobility,lin2024fedsn,hu2024accelerating,zhang2025lcfed} has emerged as a vital solution that enables collaborative model training without exposing raw data. However, as the ML model scales up, implementing FL remains a significant challenge~\cite{lyu2023optimal,lin2024adaptsfl}. For instance, training large-sized models, such as the large language model LLaMA 3-8B~\cite{zhang2025generalized} with 8 billion parameters (16GB for 16-bit floats), is computationally prohibitive for resource-constrained edge devices~\cite{fang2024automated,lin2023pushing}.


To mitigate the dilemma in FL, split learning (SL)~\cite{vepakomma2018split,lin2024split} has emerged as a promising alternative that offloads the major training workload from edge devices to a central server via layer-wise model partitioning. During model training, \zyxrev{an} edge server needs to receive smashed data (i.e., activations and gradients) from multiple edge devices for server-side training. However, as the number of edge devices increases, the massive smashed data transmission poses a substantial barrier to the deployment of SL~\cite{lin2024efficient,lin2025hierarchical}. To address this issue, recent studies~\cite{eshratifar2019bottlenet,zheng2023reducing,oh2025communication} have proposed various techniques for smashed data compression. Eshratifar~\textit{et al.}~\cite{eshratifar2019bottlenet} \zyxrev{devised} an auxiliary neural network to learn activation representations with reduced size for decreasing the communication costs between devices and the server.  Zheng~\textit{et al.}~\cite{zheng2023reducing} \zyxrev{proposed} an activation selection mechanism that retains the top-$k$ highest-magnitude elements along with a small subset of non-top-$k$ components. Oh~\textit{et al.}~\cite{oh2025communication} \zyxrev{designed} a standard deviation-based strategy to discard low-variance components and employ quantization to the remaining data for data transmission. However, these methods adopt uniform compression across all channels in the smashed data. Different channels in the smashed data encode varying types of semantic information, such as textures, edges, and object boundaries, which are closely related to the learning objective~\cite{huang2024joint}. Their contributions to model performance are not uniform--while some channels carry highly task-relevant features, others may be redundant or even introduce noise~\cite{niu2025multimodal}, thereby offering limited utility during training. Ignoring such disparities may result in suboptimal compression decisions that over-compress informative channels and under-compress redundant ones, ultimately degrading the trade-off between communication overhead and model performance.

To combat the above challenge, in this paper, we propose \zyxrev{an} \underline{SL} with \underline{a}daptive \underline{c}hannel-wise \underline{c}ompression, named SL-ACC, to significantly reduce the transmission overhead of smashed data without retaining training performance. SL-ACC consists of two key components: i) adaptive channel importance identification (ACII) and ii) channel grouping compression (CGC). Specifically, ACII quantifies the contribution of each channel in the smashed data to model training. CGC then groups the channels based on their entropy and customizes the compression \zyxrev{quantization} bit widths for each group.  The main contributions of this paper are summarized as follows.
\begin{itemize}
\item We develop ACII to quantify the contribution of each channel in the smashed data to training performance, providing the metric for adaptive smashed data compression.
\item We design CGC to group channels of the smashed data and then customize \zyxrev{quantization} bit widths for each group, so as to achieve efficient compression without compromising training accuracy.
\item We empirically evaluate SL-ACC with extensive experiments, demonstrating that the proposed SL-ACC outperforms state-of-the-art benchmarks.
\end{itemize}

The rest of the paper is organized as follows. Sec.~\ref{sec:system_design} introduces the system design of SL-ACC. Sec.~\ref{sec:performance_evaluation} provides the performance evaluation. Finally, the conclusions are provided in Sec.~\ref{sec:conclusion}.

\vspace{-0.4em}
\section{System Design}
\label{sec:system_design}
\vspace{-0.3em}
\subsection{Overview}
\vspace{-0.1em}
\begin{figure}[t]
    \centering
    \includegraphics[width=0.48\textwidth]{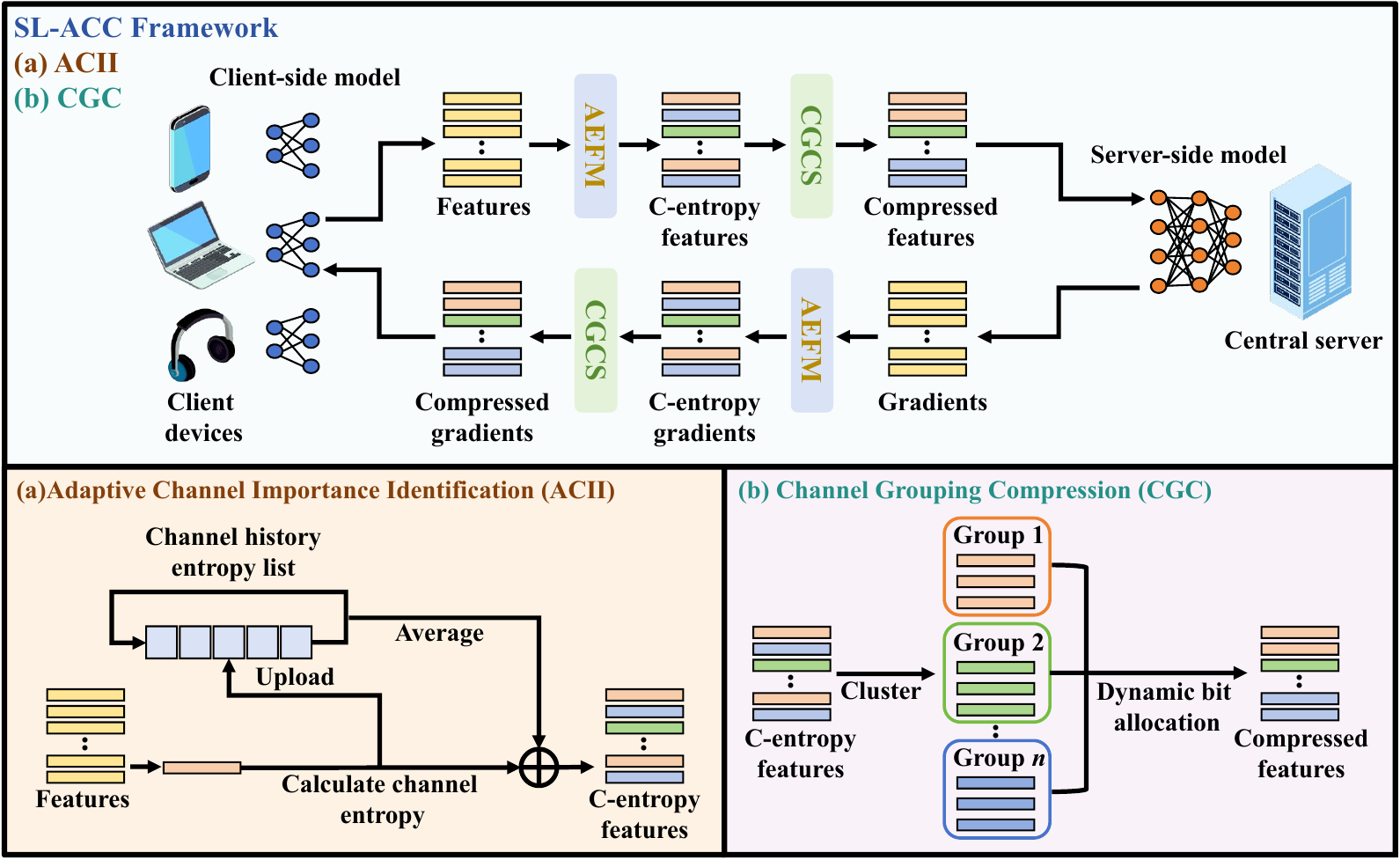} 
    \vspace{-0.6em}
    \caption{The architecture of the proposed SL-ACC framework, comprising (a) ACII for evaluating the contribution of channels in the smashed data to model training, and (b) CGC for grouping channels and customizing \zyxrev{quantization} bit widths for each group.}
    \vspace{-0.2em}
    \label{fig:aecg}
\end{figure}

In this section, we present the architecture and workflow of our proposed SL-ACC framework.  As illustrated in Fig.~\ref{fig:aecg}, SL-ACC consists of two key modules, adaptive channel importance identification (ACII) and channel grouping compression (CGC). ACII (Section~\ref{sec:aecim}) quantifies the contribution of each channel in the smashed data to model training (i.e., channel importance) based on Shannon entropy, with higher entropy indicating higher channel importance. CGC (Section~\ref{sec:CGC}) then groups channels with {similar channel importance} and allocates distinct \zyxrev{quantization} bit widths to each group.

As shown in Fig.~\ref{fig:aecg}, the global model is partitioned into client-side and server-side sub-models, deployed on the edge device and the edge server, respectively~\cite{lin2025hasfl}. The training workflow of SL-ACC in each training round comprises the following four stages: i) Each edge device executes the forward propagation of the client-side sub-model with its local dataset to generate activations; ii) The activations are processed by ACII to evaluate channel importance and then compressed by CGC. iii) The compressed activations are transmitted to the edge server, which completes the remaining forward and backward passes. The resulting gradients are again processed by ACII and CGC before being sent back; iv) Each edge device receives the compressed gradients and updates its client-side sub-model accordingly.

\vspace{-0.7em}
\subsection{Adaptive Channel Importance Identification}
\label{sec:aecim}
\vspace{-0.2em}

\begin{figure}[t]
    \centering
    \subfloat[Test accuracy.]{%
        \includegraphics[width=0.48\linewidth]{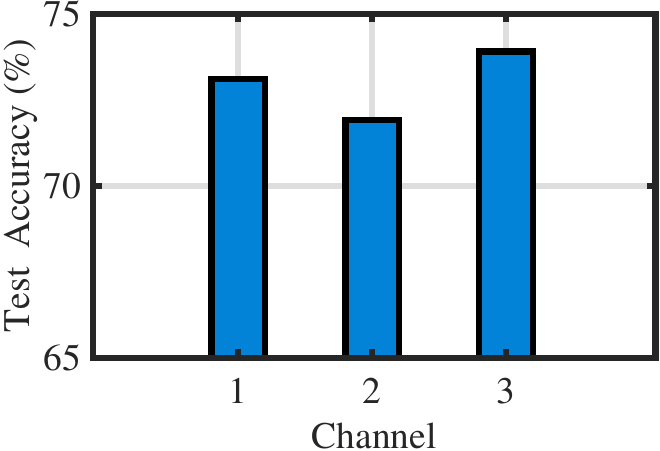}%
        \label{fig:individual_channel}
    }
    \subfloat[Dynamic accuracy changes.]{%
        \includegraphics[width=0.48\linewidth]{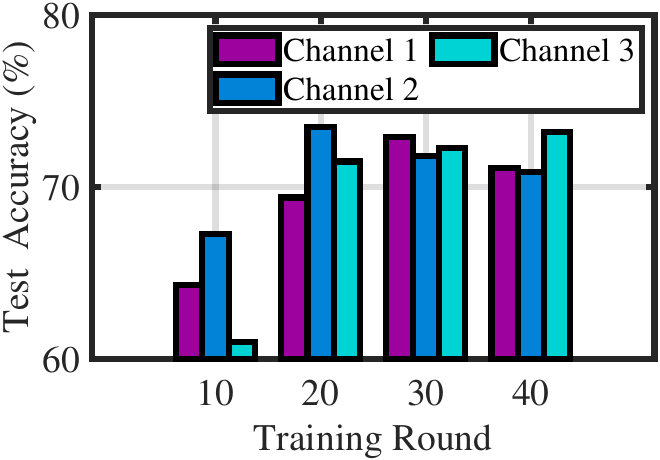}%
        \label{fig:channel_across_round}
    }
    \vspace{-0.4em}
    \caption{ The test accuracy (a) and dynamic accuracy changes (b) for individual channels, where channels 1, 2, and 3 represent different channels in smashed data.}
    \vspace{-0.6em}
    \label{fig:Channel_Information}
\end{figure}

As discussed in Sec.~\ref{sec_intro}, excessive communication overhead becomes a bottleneck as the number of participating devices increases. Given the highly uneven contributions of diverse channels to model training, a striving forward solution is to execute channel-wise compression to reduce the redundant information in the smashed data. To achieve this, it is essential to first quantify the importance of each channel to model training. To investigate the impact of individual channels in smashed data on model training, we train a model with a single channel extracted from the smashed data under the split federated learning (SFL) framework. The backbone ML model is ResNet-18, and the dataset is the HAM10000 dataset~\cite{tschandl2018ham10000}, which remains the same throughout Section~\ref{sec:system_design}. Fig.~\ref{fig:individual_channel} illustrates that individual channels contribute unequally to model performance, and Fig.~\ref{fig:channel_across_round} reveals that a channel’s contribution varies over training rounds. Therefore, it is imperative to dynamically evaluate the importance of each channel throughout the training process.

To this end, we propose ACII to quantify the importance of each channel in the smashed data for model training. Specifically, ACII first normalizes each channel’s values to the range [0,1], and then applies a softmax function to transform the normalized values into a probability distribution, where each element reflects the relative contribution of that value within the channel. To measure the information richness of each channel, entropy is \zyxrev{effective}. This is because entropy can quantify the uncertainty or disorder within a probability distribution; a higher entropy indicates a more informative distribution and thus contributes more to model training. Conversely, channels with lower entropy exhibit limited variation and contribute less to the learning process~\cite{lu2024entropy}. Therefore, we employ the entropy as an indicator of channel importance during model training. For the \( t \)-th training round, the instantaneous entropy (i.e., the entropy computed in the current training round) of the \( c \)-th channel is defined as
\begin{equation}
H_c^{(t)} = -\sum_{i=1}^{N} p_c^{(t)}(i) \log \left(p_c^{(t)}(i)\right),
\end{equation}
where \( p_c^{(t)}(i) \) denotes the normalized probability of the \( i \)-th element in the \( c \)-th channel, and \( N \) is the total number of elements in the channel (assumed equal across all channels).

While instantaneous entropy can capture the rapid changes of channel importance, it is sensitive to noise and local perturbations (e.g., randomness from mini-batch sampling), which may degrade model performance.  To overcome this limitation, we consider historical entropy, defined as the average entropy computed over multiple training rounds, in our channel importance identification. By smoothing out fluctuations over multiple training rounds, historical entropy mitigates transient noise and more accurately captures each channel’s long-term contribution to model performance.

\begin{figure}[t]
    \centering
    \subfloat[Test accuracy.]{%
        \includegraphics[width=0.48\linewidth]{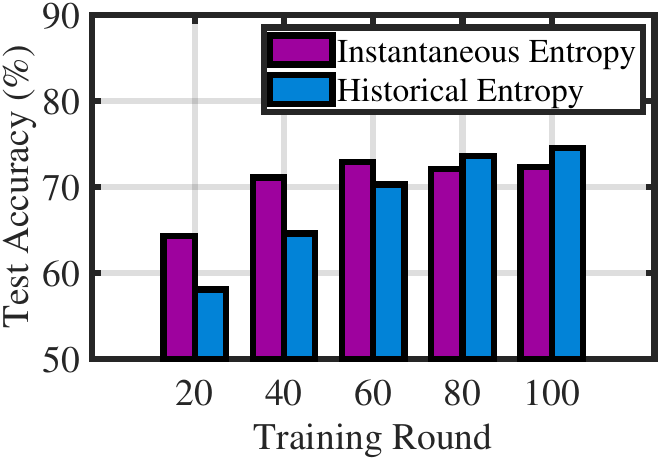}%
        \label{fig:pe21}
    }
    \subfloat[STD.]{%
        \includegraphics[width=0.48\linewidth]{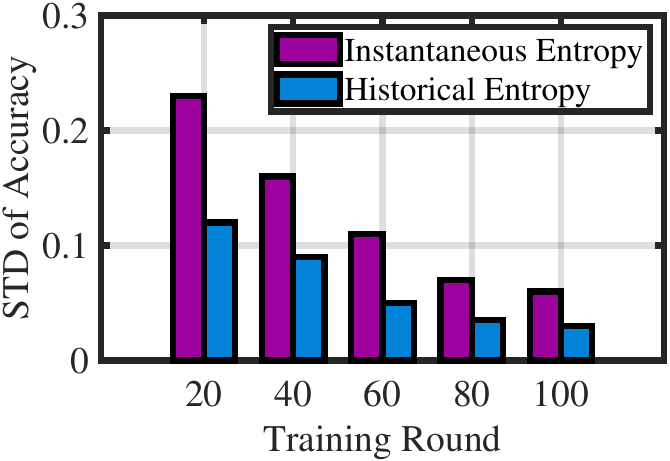}%
        \label{fig:pe22}
    }
    \vspace{-0.2em}
    \caption{The test accuracy (a) and its standard deviation (STD) (b) of the training model with the highest instantaneous and historical entropy on HAM10000 dataset.}
    \vspace{-0.2em}
    \label{fig:current_history_figures}
\end{figure}

To gain deeper insights into the impacts of instantaneous and historical entropy on model training, we select the channel with the highest instantaneous and historical entropy to train the model. Figs.~\ref{fig:pe21} and \ref{fig:pe22} show that instantaneous entropy yields faster model convergence in the early training stage but suffers from lower final accuracy and stability; in contrast, historical entropy enhances convergence stability but adapts less effectively to dynamic variations in model training. To balance this trade-off, we utilize the weighted average of the instantaneous and historical entropy to characterize the importance of the channel.  For the \( t \)-th training round, the channel entropy of the \( c \)-th channel is given by
\begin{equation}
H_c = (1 - \alpha_c^{(t)}) H_c^{(t)} + \alpha_c^{(t)} \widetilde{H}_c,
\end{equation}
where \( H_c^{(t)} \) represents the instantaneous entropy of channel \( c \) \zyxrev{and} \( \widetilde{H}_c = \frac{1}{k} \sum_{i=t-k}^{t-1} H_c^{(i)} \) denotes the historical entropy, computed as the average entropy over the past \( k \) rounds. The \zyxrev{balancing} hyperparameter \( \alpha_c^{(t)} \in [0, 1] \) controls the relative contributions of \( H_c^{(t)} \) and \( \widetilde{H}_c \).

\begin{figure}[t]
    \centering
    \subfloat[Test accuracy and time vs. \zyxrev{balancing} parameter.]{%
        \includegraphics[width=0.55\linewidth]{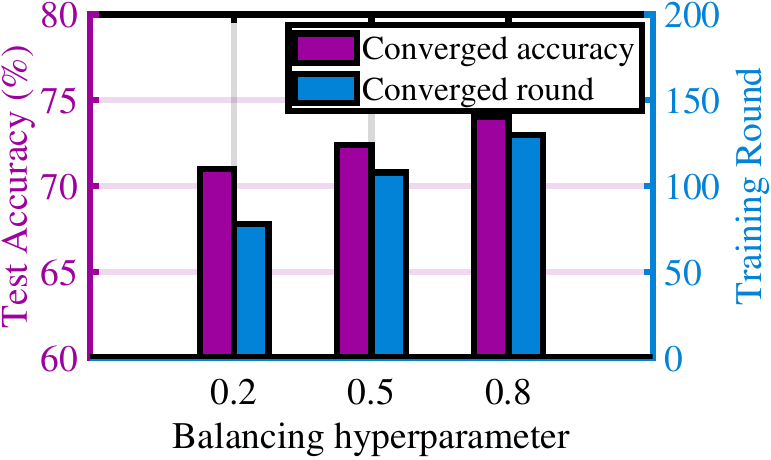}%
        \label{fig:PE31}
    }
    \subfloat[Test accuracy vs. training round]{%
        \includegraphics[width=0.47\linewidth]{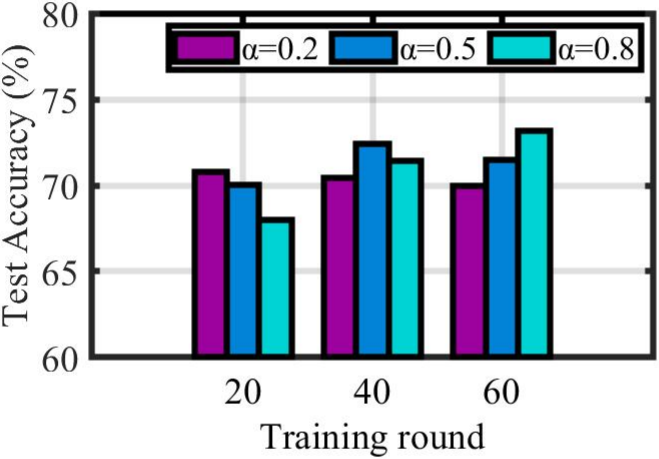}%
        \label{fig:PE32}
    }
    \vspace{-0.2em}
    \caption{The test accuracy and time versus \zyxrev{balancing} hyperparameter (a) and test accuracy across 60 training rounds with varying \zyxrev{balancing} hyperparameter (b) on HAM10000 dataset.}
\vspace{-0.4em}
    \label{fig:PE3}
\end{figure}

The key to identifying the channel importance lies in the design of the \zyxrev{balancing} hyperparameter \(\alpha_c^{(t)}\). To investigate its impact on training performance, we conduct experiments with varying \zyxrev{balancing} hyperparameters. Fig.~\ref{fig:PE31} illustrates that \zyxrev{balancing} hyperparameter trade off convergence speed and training accuracy, while Fig.~\ref{fig:PE32} demonstrates that the optimal \zyxrev{balancing} hyperparameter varies across training rounds. These observations motivate the need for the dynamic adjustment of the \zyxrev{balancing} hyperparameter throughout training. Thus, we design the \zyxrev{balancing} hyperparameter $\alpha_c^{(t)}$ as a function of the training stage. In the early training stage, the model benefits from emphasizing instantaneous entropy, which captures current training dynamics to expedite convergence. \zyxrev{A}s training progresses, over-reliance on instantaneous entropy could overfit to noise and local fluctuations, thereby degrading the training performance. Therefore, the model should gradually shift its focus to historical entropy. We control this transition by using the ratio of the current round index $t$ to the total number of training rounds $T$, as follows
\begin{equation}
\alpha_c^{(t)} = \frac{t}{T}.
\end{equation}

\vspace{-0.6em}
\subsection{Channel Grouping Compression}
\label{sec:CGC}

To tailor communication-efficient distributed ML, quantization-based compression is widely used to convert high-precision data into low-bit representations with minimal performance loss~\cite{oh2025communication}. However, existing methods~\cite{eshratifar2019bottlenet,zheng2023reducing,oh2025communication} typically employ uniform compression across all channels in the smashed data, ignoring variations in their importance to model training. This often results in over-compressing high-importance channels and under-compressing low-importance ones, ultimately degrading overall model performance.

To address this issue, we propose CGC, which groups the channels of smashed data based on their channel entropy shown in Section~\ref{sec:aecim}, and then allocates different quantization bit widths to each group. Specifically, CGC first clusters channels with similar channel entropy $H_c$ into a group. Given the low-dimensional and continuous nature of the entropy space, where Euclidean distance reliably captures similarity, K-means~\cite{zhao2022adaptive}, which is good at handling low-dimensional data, offers a well-suited solution. By leveraging the principle of K-means, we can divide the channels into \(g\) disjoint groups $\{\mathcal{G}_1,\mathcal{G}_2, \dots, \mathcal{G}_g\}$ via solving the following optimization problem
\begin{equation}
\{\mathcal{G}_1,\mathcal{G}_2, \dots, \mathcal{G}_g\} =
\underset{\mathcal{P}}{\arg\min}
\sum_{j=1}^{g} \sum_{c=1}^{m_j} \left\| H_{j,c} - \mu_j \right\|^2,
\end{equation}
where $\mathcal{G}_j$ is the $j$-th group; \( H_{j,c} \), \( \mu_j \),  and \( m_j \) denote the entropy of the \( c \)-th channel,  mean entropy, and the number of channels in $\mathcal{G}_j$, respectively; \( g \) is the total number of groups; \( \mathcal{P} \) denotes the set of all possible of the channels into \( g \) disjoint groups; \( \underset{\mathcal{X}}{\arg\min} f(\cdot) \) returns the argument within the domain \(\mathcal{X}\) that minimizes the objective function \(f(\cdot)\).

After channel grouping, we can compute the average entropy of each group as an indicator of its contribution to training. For a given group $\mathcal{G}_j$, the average entropy \( \widetilde{H}_j \) is expressed as 
\begin{equation}
\widetilde{H}_j = \frac{1}{m_j} \sum_{c=1}^{m_j} H_{j,c}.
\label{eq:avg_entropy}
\end{equation}

The average entropy of a group quantifies its overall information richness, where higher entropy indicates that the group contains more informative channels, which require more bits to reduce quantization-induced information loss. Conversely, a lower entropy suggests that the channels are less informative and can be quantized with fewer bits to enhance compression efficiency. In light of this, we adaptively customize the quantization bit width \( b_j \) to each channel within the group \(\mathcal{G}_j\) based on its average entropy \( \widetilde{H}_j \). For the $j$-th group, the allocated quantization bit width is given by
\begin{equation}
b_j = \min\left(b_{\max}, \max\left(b_{\min}, \left\lfloor \widetilde{H}_j \right\rfloor\right)\right),
\end{equation}
where \( b_{\min} \) and \( b_{\max} \) denote the lower and upper bounds of the quantization bit width, respectively. 

Since channels in the same group exhibit not only similar levels of information richness but also homogeneous statistical distributions, particularly in value ranges and variance~\cite{zhong2020channel}. Linear quantization assumes that the underlying data is approximately uniformly distributed within a bounded range. While this assumption may not hold globally across all channels, it becomes more valid locally within each clustered group, where values are empirically observed to concentrate within a relatively narrow and consistent range. This renders linear quantization an effective and low-error approximation method for these distributions~\cite{moon2024instance}.
Thus, CGC applies the linear quantization with boundaries of the minimum value \( x_{j,\min} \) and the maximum value \( x_{j,\max} \), to each channel group. The quantization procedure is shown as follows
\begin{equation}
\hat{x} = \text{round}\left( \frac{x - x_{j,\min}}{x_{j,\max} - x_{j,\min}} \cdot (2^{b_j} - 1) \right),
\end{equation}
where \text{round}($\cdot$) denotes the rounding function that returns the nearest integer to its input, with half-integer values rounded away from zero to reduce quantization bias.

\vspace{-0.3em}
\section{Performance Evaluation}
\label{sec:performance_evaluation}
In this section, we introduce the experimental setup and evaluate SL-ACC against state-of-the-art benchmarks.
\vspace{-0.2em}

\begin{figure}[t]
    \centering
    \subfloat[\rmfamily\footnotesize HAM10000 (IID).]{%
        \includegraphics[width=0.48\linewidth]{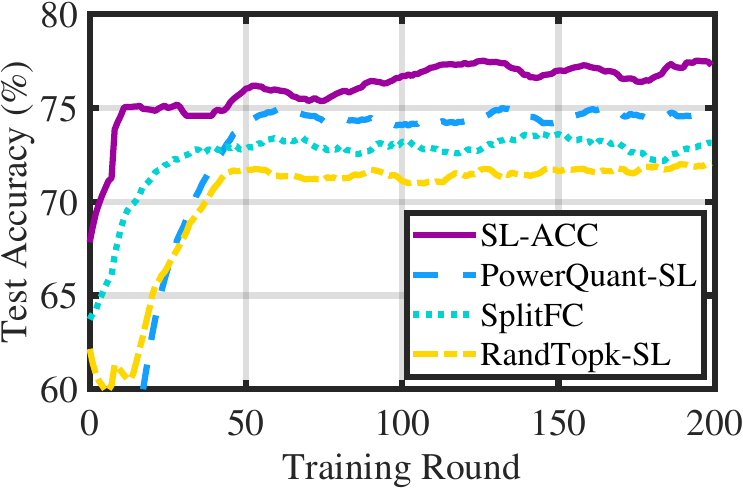}%
        \label{fig:Main_HM100001}
    }
    \subfloat[\rmfamily\footnotesize HAM10000 (non-IID).]{%
        \includegraphics[width=0.48\linewidth]{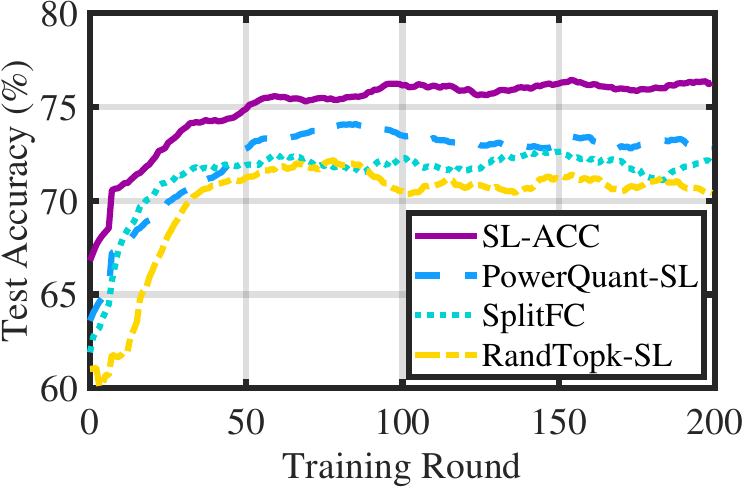}%
        \label{fig:Main_HM100002}
    }\\[-1.0ex]
    \subfloat[\rmfamily\footnotesize MNIST (IID).]{%
        \includegraphics[width=0.48\linewidth]{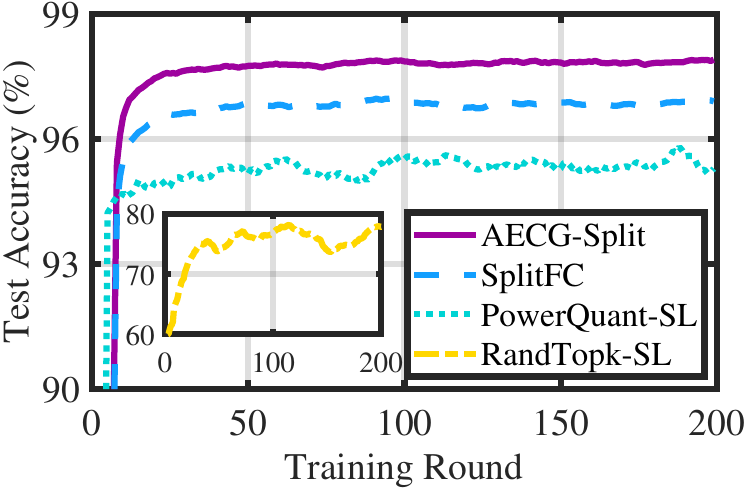}%
        \label{fig:Main_MNIST1}
    }
    \subfloat[\rmfamily\footnotesize MNIST (non-IID).]{%
        \includegraphics[width=0.48\linewidth]{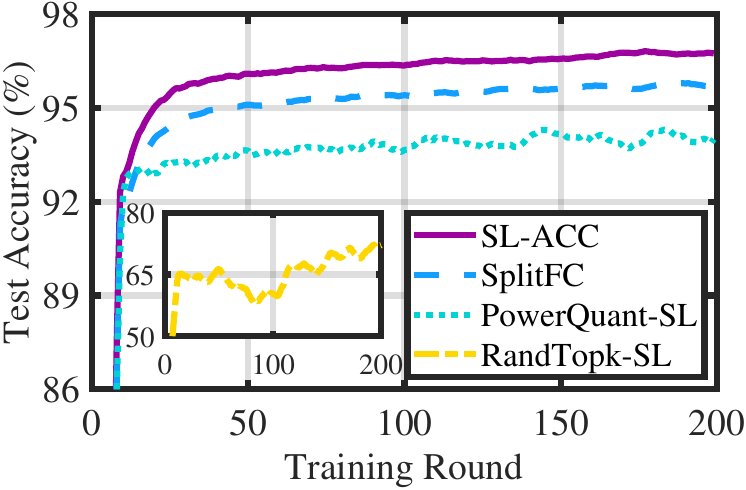}%
        \label{fig:Main_MNIST2}
    }
    \vspace{-0.2em}
    \caption{\rmfamily The training performance of ResNet-18 on the HAM10000 and MNIST datasets under both IID and non-IID settings.}
    \vspace{-0.4em}
    \label{fig:Superiority_SLACC}
\end{figure}

\vspace{-0.99em}
\subsection{Experiment Setup}

\subsubsection{Implementation}
The experiments are conducted on two high-performance servers, each equipped with six NVIDIA RTX 3090 GPUs. To emulate the SL architecture, one server is utilized to emulate multiple edge devices, while the other serves as the edge server. The client-side and server-side sub-models training are executed in parallel using distributed data parallelism to ensure efficient execution. The software stack includes Python 3.7.2 and PyTorch 1.2.0.

\subsubsection{Dataset and Model}
We evaluate the performance of SL-ACC on two classification datasets: HAM10000~\cite{tschandl2018ham10000} and MNIST~\cite{lecun1998gradient}. 
Experiments are conducted under both (independent and identically distributed) IID and non-IID settings. For the IID setting, data samples are randomly shuffled and evenly distributed across devices. For the non-IID setting, data is partitioned using a Dirichlet distribution with $\beta = 0.5$. To implement SL-ACC, we employ the ResNet-18 network as the global model, where the first three layers are deployed on edge devices as the client-side sub-model, and the remaining layers reside on the server as the server-side sub-model. 

\subsubsection{Benchmarks}


To comprehensively evaluate the performance of SL-ACC, we compare it with the following alternatives: i) \textbf{PowerQuant-SL} is the SL variant of PowerQuant~\cite{yvinec2023powerquant}, which utilizes a power function to quantize the smashed data for compression; ii) \textbf{RandTopk-SL} is the SL variant of top-$k$~\cite{zheng2023reducing}, which compresses the smashed data by retaining the top-$k$ elements with the highest magnitudes along with a small subset of randomly selected non-top-$k$ elements; iii) \textbf{SplitFC}~\cite{oh2025communication} compresses the smashed data by discarding low-variance features based on standard deviation, followed by quantization of the remaining elements.





\subsubsection{Hyperparameters}

In our experiments, we utilize a stochastic gradient descent optimizer. The learning rate is set to 0.0001, and the mini-batch size is set to 128. The lower and upper bounds of the quantization bit width are set to 2 and 8, respectively. Unless otherwise specified, the number of edge devices is set to 5.

\vspace{-0.8em}
\subsection{Superiority of SL-ACC}
\vspace{-0.2em}
Fig.~\ref{fig:Superiority_SLACC} shows that SL-ACC consistently outperforms all benchmarks on both HAM10000 and MNIST under IID and non-IID settings. SL-ACC achieves 78.91\% and 76.83\% on HAM10000, and 98.15\% and 97.09\% on MNIST, under IID and non-IID conditions, respectively. The reason is twofold: One is that ACII effectively identifies channels by considering both instantaneous and historical importance, and the other is that CGC preserves channel information by assigning higher quantization bit widths to high-importance channels and reducing information redundancy by allocating lower quantization bit widths to low-importance channels. In contrast, SplitFC achieves a test accuracy of 74.16\% on HAM10000 under the IID setting, which is 4.75\% lower than that of SL-ACC. This performance degradation is primarily due to its reliance on standard deviation-based feature selection, which is sensitive to noise and often discards low-variance yet informative channels. Similarly, PowerQuant-SL and RandTopk-SL achieve 73.96\% and 71.56\% on HAM10000 and MNIST under non-IID settings, which are 2.87\% and 25.53\% lower than those of SL-ACC. Their performance decline results from compression strategies that ignore the contribution of the channel to model training.




\vspace{-0.7em}

\subsection{Ablation Study}
\subsubsection{Adaptive Channel Importance Identification (ACII)}

\begin{figure}[t]
    \centering
    \subfloat[\rmfamily\footnotesize ACII (IID).]{%
        \includegraphics[width=0.48\linewidth]{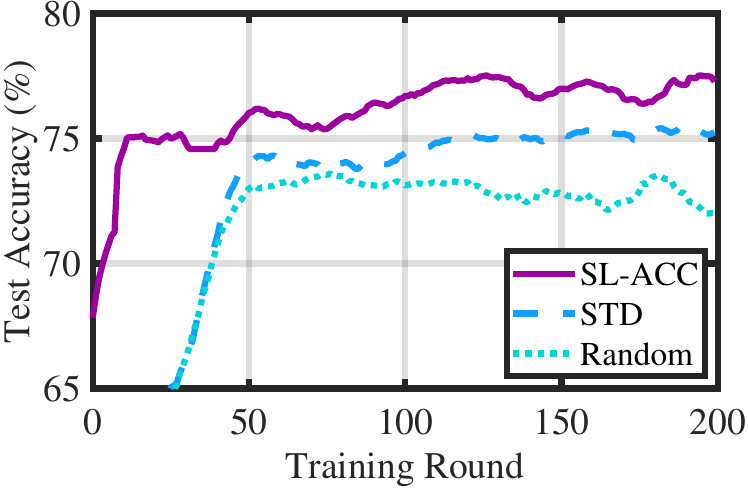}%
        \label{fig:pe1}
    }
    \subfloat[\rmfamily\footnotesize ACII (non-IID).]{%
    \includegraphics[width=0.48\linewidth]{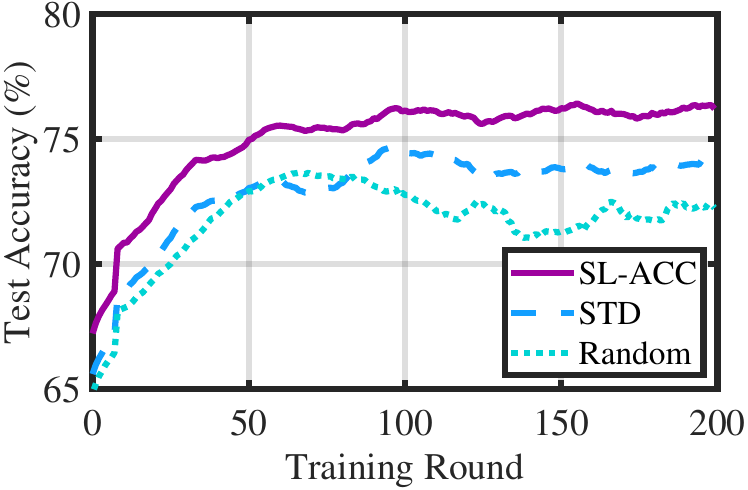}%
    \label{fig:pe1}
    }
    \vspace{-0.2em}
    \caption{\rmfamily The ablation experiments for ACII on the HAM10000 dataset under IID and non-IID settings.}
    \label{fig:ablation_AECIM}
    \vspace{-0.8em}
\end{figure}

Fig.~\ref{fig:ablation_AECIM} illustrates the impact of ACII on training performance by comparing it with random- and STD-based channel selection (i.e., selecting a channel with the highest STD). SL-ACC achieves faster convergence and higher test accuracy, outperforming STD- and Random-based selection by 3.79\% (2.21\%) and 5.24\% (3.29\%) under the IID (non-IID) setting, respectively. This improvement stems from ACII’s use of both instantaneous and historical entropy to identify training-critical channels. In contrast, random- and STD-based selection introduce more noise and fail to prioritize important channels, resulting in slower convergence and lower accuracy.

\subsubsection{Channel Grouping Compression (CGC)}
\begin{figure}[t]
    \centering
    \subfloat[\rmfamily\footnotesize CGC (IID).]{%
        \includegraphics[width=0.48\linewidth]{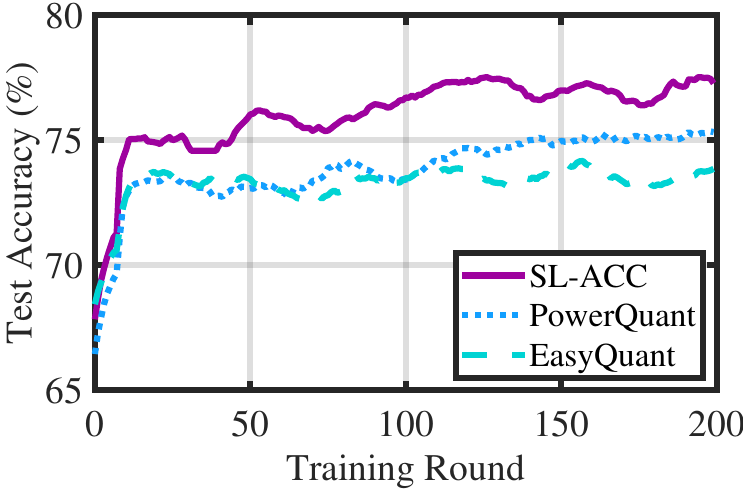}%
        \label{fig:pe12}
    }
    \subfloat[\rmfamily\footnotesize CGC (non-IID).]{%
    \includegraphics[width=0.48\linewidth]{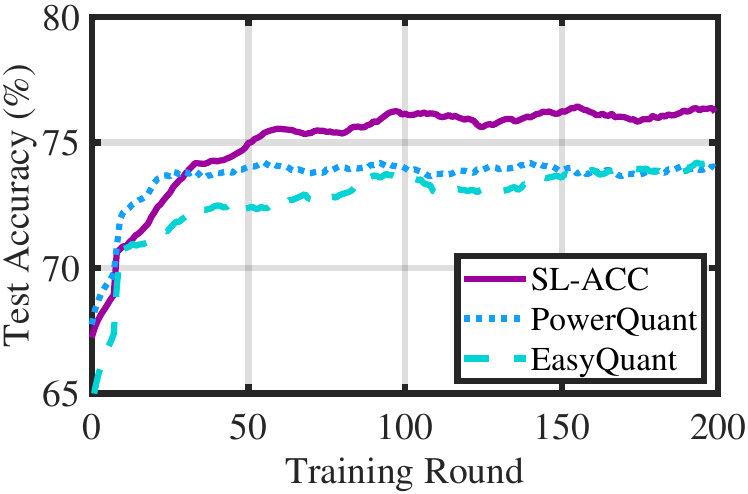}%
    \label{fig:pe12}
    }
    \vspace{-0.2em}
    \caption{\rmfamily The ablation experiments for CGC on the HAM10000 dataset under IID and non-IID settings.}
    \vspace{-0.8em}
    \label{fig:ablation_CGC}
\end{figure}

Fig.~\ref{fig:ablation_CGC} illustrates the impact of CGC on model performance by comparing it with PowerQuant~\cite{yvinec2023powerquant} and EasyQuant~\cite{tang2023easyquant}. SL-ACC achieves convergence accuracy of 78.91\% (76.83\%), surpassing EasyQuant and PowerQuant by 4.88\% (2.67\%) and 3.71\% (3.19\%) under the IID (non-IID) setting. While the benchmarks employ uniform scaling or power-law scaling, they adopt fixed bit-width allocation across all channels. This often results in over-compressing high-importance channels and under-compressing low-importance ones, ultimately degrading overall model performance. In contrast, CGC groups channels based on their importance to model training and then assigns bit-widths for each group, reducing communication cost without compromising training performance.

\vspace{-0.6em}
\section{Conclusion}
\label{sec:conclusion}
\vspace{-0.2em}
In this paper, we have proposed a communication-efficient SL framework, named SL-ACC, aimed at reducing communication overhead in SL without compromising training performance. SL-ACC consists of two key components: ACII and CGC. First, ACII leverages Shannon entropy to quantify the contribution of each channel in the smashed data to model training. Second, CGC clusters channels based on their entropy and conducts adaptive compression for each group, reducing data transmission redundancy while preserving model performance. Extensive experiments across multiple datasets demonstrate that SL-ACC consistently outperforms the state-of-the-art benchmarks.


\vspace{-1.2em}
{
    \bibliographystyle{IEEEtran}
    \bibliography{references}

\begin{thebibliography}{10}
\providecommand{\url}[1]{#1}
\csname url@samestyle\endcsname
\providecommand{\newblock}{\relax}
\providecommand{\bibinfo}[2]{#2}
\providecommand{\BIBentrySTDinterwordspacing}{\spaceskip=0pt\relax}
\providecommand{\BIBentryALTinterwordstretchfactor}{4}
\providecommand{\BIBentryALTinterwordspacing}{\spaceskip=\fontdimen2\font plus
\BIBentryALTinterwordstretchfactor\fontdimen3\font minus \fontdimen4\font\relax}
\providecommand{\BIBforeignlanguage}[2]{{%
\expandafter\ifx\csname l@#1\endcsname\relax
\typeout{** WARNING: IEEEtran.bst: No hyphenation pattern has been}%
\typeout{** loaded for the language `#1'. Using the pattern for}%
\typeout{** the default language instead.}%
\else
\language=\csname l@#1\endcsname
\fi
#2}}
\providecommand{\BIBdecl}{\relax}
\BIBdecl

\bibitem{rydning2023worldwide}
J.~Rydning, ``{Worldwide IDC Global Datasphere Forecast, 2023--2027: It's a Distributed, Diverse, and Dynamic (3D) Datasphere},'' \emph{IDC Market Forecast}, no. US50554523, Dec. 2023.

\bibitem{lin2022channel}
Z.~Lin, L.~Wang, J.~Ding, B.~Tan, and S.~Jin, ``{Channel Power Gain Estimation for Terahertz Vehicle-to-Infrastructure Networks},'' \emph{{IEEE} Commun. Lett.}, vol.~27, no.~1, pp. 155--159, 2022.

\bibitem{wang2018networking}
J.~Wang, J.~Liu, and N.~Kato, ``Networking and communications in autonomous driving: A survey,'' \emph{IEEE Communications Surveys \& Tutorials}, vol.~21, no.~2, pp. 1243--1274, 2018.

\bibitem{lin2022tracking}
Z.~Lin, L.~Wang, J.~Ding, Y.~Xu, and B.~Tan, ``Tracking and transmission design in terahertz v2i networks,'' \emph{IEEE Transactions on Wireless Communications}, vol.~22, no.~6, pp. 3586--3598, 2022.

\bibitem{fang2025dynamic}
Z.~Fang, Z.~Lin, S.~Hu, Y.~Tao, Y.~Deng, X.~Chen, and Y.~Fang, ``Dynamic uncertainty-aware multimodal fusion for outdoor health monitoring,'' \emph{arXiv preprint arXiv:2508.09085}, 2025.

\bibitem{tang2024merit}
Y.~Tang, Z.~Chen, A.~Li, T.~Zheng, Z.~Lin, J.~Xu, P.~Lv, Z.~Sun, and Y.~Gao, ``{MERIT: Multimodal Wearable Vital Sign Waveform Monitoring},'' \emph{arXiv preprint arXiv:2410.00392}, 2024.

\bibitem{fang2024ic3m}
Z.~Fang, Z.~Lin, S.~Hu, H.~Cao, Y.~Deng, X.~Chen, and Y.~Fang, ``{IC3M: In-Car Multimodal Multi-Object Monitoring for Abnormal Status of Both Driver and Passengers},'' \emph{arXiv preprint arXiv:2410.02592}, 2024.

\bibitem{qiu2024ifvit}
Y.~Qiu, H.~Chen, X.~Dong, Z.~Lin, I.~Y. Liao, M.~Tistarelli, and Z.~Jin, ``Ifvit: Interpretable fixed-length representation for fingerprint matching via vision transformer,'' \emph{IEEE Transactions on Information Forensics and Security}, 2024.

\bibitem{lin2025hsplitlora}
Z.~Lin, Y.~Zhang, Z.~Chen, Z.~Fang, X.~Chen, P.~Vepakomma, W.~Ni, J.~Luo, and Y.~Gao, ``{HSplitLoRA: A Heterogeneous Split Parameter-Efficient Fine-Tuning Framework for Large Language Models},'' \emph{arXiv preprint arXiv:2505.02795}, 2025.

\bibitem{sai2024dmdat}
S.~Sai, U.~Mittal, and V.~Chamola, ``{DMDAT: Diffusion Model-based Data Augmentation Technique for Vision-based Accident Detection in Vehicular Networks},'' \emph{IEEE Trans. Veh. Technol.}, vol.~74, no.~2, pp. 2241--2250, Feb. 2025.

\bibitem{chen2025mobility}
T.~Chen, J.~Yan, Y.~Sun, S.~Zhou, D.~G{\"u}nd{\"u}z, and Z.~Niu, ``{Mobility Accelerates Learning: Convergence Analysis on Hierarchical Federated Learning in Vehicular Networks},'' \emph{IEEE Trans. Veh. Technol.}, vol.~74, no.~1, pp. 1657--1673, Jan. 2025.

\bibitem{lin2024fedsn}
Z.~Lin, Z.~Chen, Z.~Fang, X.~Chen, X.~Wang, and Y.~Gao, ``Fedsn: A federated learning framework over heterogeneous leo satellite networks,'' \emph{IEEE Transactions on Mobile Computing}, 2024.

\bibitem{hu2024accelerating}
M.~Hu, J.~Zhang, X.~Wang, S.~Liu, and Z.~Lin, ``{Accelerating Federated Learning with Model Segmentation for Edge Networks},'' \emph{{IEEE} Trans. Green Commun. Netw.}, 2024.

\bibitem{zhang2025lcfed}
Y.~Zhang, H.~Chen, Z.~Lin, Z.~Chen, and J.~Zhao, ``{LCFed: An Efficient Clustered Federated Learning Framework for Heterogeneous Data},'' \emph{arXiv preprint arXiv:2501.01850}, 2025.

\bibitem{lyu2023optimal}
S.~Lyu, Z.~Lin, G.~Qu, X.~Chen, X.~Huang, and P.~Li, ``Optimal resource allocation for u-shaped parallel split learning,'' in \emph{2023 IEEE Globecom Workshops (GC Wkshps)}, 2023, pp. 197--202.

\bibitem{lin2024adaptsfl}
Z.~Lin, G.~Qu, W.~Wei, X.~Chen, and K.~K. Leung, ``{Adaptsfl: Adaptive Split Federated Learning in Resource-Constrained Edge Networks},'' \emph{{IEEE} Trans. Netw.}, 2024.

\bibitem{zhang2025generalized}
H.~Zhang, C.~Hou, J.~Chen, H.~Zhang, and F.-Y. Wang, ``{A Generalized ChatGPT-based Collaborative Multi-objective Decision-making Framework for Robust Vehicle Platoon Collision Avoidance},'' \emph{IEEE Trans. Veh. Technol.}, vol.~74, no.~5, pp. 7212--7225, May 2025.

\bibitem{fang2024automated}
Z.~Fang, Z.~Lin, Z.~Chen, X.~Chen, Y.~Gao, and Y.~Fang, ``{Automated Federated Pipeline for Parameter-Efficient Fine-Tuning of Large Language Models},'' \emph{arXiv preprint arXiv:2404.06448}, 2024.

\bibitem{lin2023pushing}
Z.~Lin, G.~Qu, Q.~Chen, X.~Chen, Z.~Chen, and K.~Huang, ``{Pushing Large Language Models to the 6G Edge: Vision, Challenges, and Opportunities},'' \emph{IEEE Communications Magazine}, 2025.

\bibitem{vepakomma2018split}
P.~Vepakomma, O.~Gupta, T.~Swedish, and R.~Raskar, ``{Split Learning for Health: Distributed Deep Learning Without Sharing Raw Patient Data},'' in \emph{Proc. of the 7th ICLR Workshop}, 2019.

\bibitem{lin2024split}
Z.~Lin, G.~Qu, X.~Chen, and K.~Huang, ``{Split Learning in 6G Edge Networks},'' \emph{{IEEE} Wirel. Commun.}, 2024.

\bibitem{lin2024efficient}
Z.~Lin, G.~Zhu, Y.~Deng, X.~Chen, Y.~Gao, K.~Huang, and Y.~Fang, ``{Efficient Parallel Split Learning over Resource-Constrained Wireless Edge Networks},'' \emph{{IEEE} Trans. Mobile Comput.}, vol.~23, no.~10, pp. 9224--9239, 2024.

\bibitem{lin2025hierarchical}
Z.~Lin, W.~Wei, Z.~Chen, C.-T. Lam, X.~Chen, Y.~Gao, and J.~Luo, ``{Hierarchical Split Federated Learning: Convergence Analysis and System Optimization},'' \emph{{IEEE} Trans. Mobile Comput.}, 2025.

\bibitem{eshratifar2019bottlenet}
A.~E. Eshratifar, A.~Esmaili, and M.~Pedram, ``{BottleNet: A Deep Learning Architecture for Intelligent Mobile Cloud Computing Services},'' in \emph{Proc. of the 25th ISLPED}, Jul. 2019, pp. 1--6.

\bibitem{zheng2023reducing}
F.~Zheng, C.~Chen, L.~Lyu, and B.~Yao, ``{Reducing Communication for Split Learning by Randomized Top-k Sparsification},'' in \emph{Proc. of the 32nd IJCAI}, 2023, pp. 4665--4673.

\bibitem{oh2025communication}
Y.~Oh, J.~Lee, C.~G. Brinton, and Y.-S. Jeon, ``{Communication-Efficient Split Learning via Adaptive Feature-Wise Compression},'' \emph{IEEE Trans. Neural Netw. Learn. Syst.}, vol.~36, no.~6, pp. 10\,844--10\,858, Jan. 2025.

\bibitem{huang2024joint}
Z.~Huang, Y.~Jia, Y.~Zhang, X.~Liu, H.~Shen, and W.~Wen, ``{Joint Source-channel Coding for Image Super-resolution Tasks in Semantic Communications},'' \emph{IEEE Trans. Veh. Technol.}, vol.~73, no.~12, pp. 19\,034--19\,039, Dec. 2024.

\bibitem{niu2025multimodal}
X.~Niu, L.~Tan, J.~Wu, W.~Yuan, and T.~Q. Quek, ``{Multimodal-oriented Interactive Joint Source-channel Coding for Lightweight Semantic Communication},'' \emph{IEEE Trans. Veh. Technol.}, May 2025, early access.

\bibitem{lin2025hasfl}
Z.~Lin, Z.~Chen, X.~Chen, W.~Ni, and Y.~Gao, ``{HASFL: Heterogeneity-Aware Split Federated Learning over Edge Computing Systems},'' \emph{arXiv preprint arXiv:2506.08426}, 2025.

\bibitem{tschandl2018ham10000}
P.~Tschandl, C.~Rosendahl, and H.~Kittler, ``{The HAM10000 Dataset, A Large Collection of Multi-source Dermatoscopic Images of Common Pigmented Skin Lesions},'' \emph{Sci. Data}, vol.~5, no.~1, pp. 1--9, Aug. 2018.

\bibitem{lu2024entropy}
Y.~Lu, Z.~Guan, Y.~Yang, W.~Zhao, M.~Gong, and C.~Xu, ``{Entropy Induced Pruning Framework for Convolutional Neural Networks},'' in \emph{Proc. of the 38th AAAI}, 2024, pp. 3918--3926.

\bibitem{zhao2022adaptive}
H.~Zhao, J.~Tang, B.~Adebisi, T.~Ohtsuki, G.~Gui, and H.~Zhu, ``{An Adaptive Vehicle Clustering Algorithm Based on Power Minimization in Vehicular Ad-hoc Networks},'' \emph{IEEE Trans. Veh. Technol.}, vol.~71, no.~3, pp. 2939--2948, Mar. 2022.

\bibitem{zhong2020channel}
Z.~Zhong, H.~Akutsu, and K.~Aizawa, ``{Channel-Level Variable Quantization Network for Deep Image Compression},'' in \emph{Proc. of the 29nd IJCAI}, 2020, pp. 467--473.

\bibitem{moon2024instance}
J.~Moon, D.~Kim, J.~Cheon, and B.~Ham, ``{Instance-aware Group Quantization for Vision Transformers},'' in \emph{Proc. of the 29th IEEE/CVF CVPR}, 2024, pp. 16\,132--16\,141.

\bibitem{lecun1998gradient}
Y.~LeCun, L.~Bottou, Y.~Bengio, and P.~Haffner, ``{Gradient-based Learning Applied to Document Recognition},'' \emph{Proc. IEEE}, vol.~86, no.~11, pp. 2278--2324, Nov. 1998.

\bibitem{yvinec2023powerquant}
E.~Yvinec, A.~Dapogny, M.~Cord, and K.~Bailly, ``{PowerQuant: Automorphism Search for Non-Uniform Quantization},'' in \emph{Proc. of the 11nd ICLR}, 2023, pp. 1--21.

\bibitem{tang2023easyquant}
H.~Tang, Y.~Bai, Y.~Wang, Z.~Wang, and Y.~Wang, ``{EasyQuant: An Efficient Data-free Quantization Algorithm for LLMs},'' in \emph{Proc. of the 29th EMNLP}, 2023, pp. 9119--9128.

\end{thebibliography}
}
\vspace{-1.2em}
\end{document}